\pdfoutput=1

\documentclass[11pt]{article}

\usepackage{emnlp2021}
\usepackage{times}
\usepackage{latexsym}

\usepackage{graphicx}

\usepackage[T1]{fontenc}

\usepackage[utf8]{inputenc}

\usepackage{microtype}

\usepackage{booktabs}
\usepackage{bm}
\usepackage{multirow}

\usepackage{xspace}
\def\SAS{SAS\xspace}

\newcommand{\B}{\bfseries}

\newcommand*\samethanks[1][\value{footnote}]{\footnotemark[#1]}

\title{Semantic Answer Similarity for Evaluating Question Answering Models}

\author{Julian Risch\thanks{\ \ Both authors contributed equally to this research.} \and Timo Möller\samethanks \and Julian Gutsch \and Malte Pietsch\\
         deepset\\ \texttt{\small \{julian.risch, timo.moeller, julian.gutsch, malte.pietsch\}@deepset.ai}}

\begin{document}
\maketitle
\begin{abstract}
The evaluation of question answering models compares ground-truth annotations with model predictions.
However, as of today, this comparison is mostly lexical-based and therefore misses out on answers that have no lexical overlap but are still semantically similar, thus treating correct answers as false. 
This underestimation of the true performance of models hinders user acceptance in applications and complicates a fair comparison of different models.
Therefore, there is a need for an evaluation metric that is based on semantics instead of pure string similarity. 
In this short paper, we present \SAS, a cross-encoder-based metric for the estimation of semantic answer similarity, and compare it to seven existing metrics.
To this end, we create an English and a German three-way annotated evaluation dataset containing pairs of answers along with human judgment of their semantic similarity, which we release along with an implementation of the \SAS metric and the experiments. 
We find that semantic similarity metrics based on recent transformer models correlate much better with human judgment than traditional lexical similarity metrics on our two newly created datasets and one dataset from related work.
\end{abstract}

\section{Introduction}
The evaluation of question answering (QA) models relies on human-annotated datasets of question-answer pairs.
Given a question, the ground-truth answer is compared to the answer predicted by a model with regard to different similarity metrics.
Currently, the most prominent metrics for the evaluation of QA models are exact match (EM), F1-score, and top-n-accuracy.
All these three metrics rely on string-based comparison.
EM is a binary metric that checks whether the predicted answer string matches exactly the ground-truth answer.
While this metric works well for short factual answers, such as names of persons or locations, it has some obvious flaws when it comes to comparing slightly differing short answers or longer, more elaborate answers.
Even a prediction that differs from the ground truth in only one character in the answer string is evaluated as completely wrong.
To mitigate this problem and have a continuous score ranging between 0 and 1, the F1-score can be used.
In this case, precision is calculated based on the relative number of tokens in the prediction that are also in the ground-truth answer and recall is calculated based on the relative number of tokens in the ground-truth answer that are also in the prediction.
An extension of this metric runs stop word removal and lowercasing before the comparison, for example, to disregard prepositions.
As an F1-score is not as simple to interpret as accuracy, there is a third common metric for the evaluation of QA models.
Top-n-accuracy evaluates the first n model predictions as a group and considers the predictions correct if there is any \emph{positional} overlap between the ground-truth answer and one of the first n model predictions --- otherwise, they are considered incorrect. 
The answer string itself is not compared for top-n-accuracy but the start and end index of the answer within the text document from where the answer is extracted, called context.

If a dataset contains multi-way annotations, there can be multiple different ground-truth answers for the same question.
The maximum similarity score of a prediction over all ground-truth answers is used in that case, which works with all the metrics above.
However, a problem is that sometimes only one correct answer is annotated when in fact there are multiple correct answers in a document.
If the multiple correct answers are semantically but not lexically the same, existing metrics require all correct answers within a document to be annotated and cannot be used reliably otherwise.
%
Figure~\ref{fig:placeholder2} gives an example of a context, a question, multiple ground-truth answers, a prediction and different similarity scores.
Besides EM, F1-score, and top-1-accuracy, we also list human judgment.
The example shows that the existing metrics cannot capture the semantic similarity of the prediction and the ground-truth answers but are limited to lexical similarity.
\setlength{\fboxsep}{8pt}
\begin{figure}
\centering
\fbox{\parbox{0.90\linewidth}{
\textbf{Question:} How many plant species are estimated to be in the Amazon region?\\
\textbf{Context:} 
The region is home to about 2.5 million insect species, tens of thousands of plants, and some 2,000 birds and mammals. To date, at least 40,000 plant species [\ldots]\\
\textbf{Ground-Truth Answer:} ``40,000''\\
\textbf{Predicted Answer:} ``tens of thousands''\\
\textbf{Exact Match:} 0.00\\
\textbf{F1-Score:} 0.00 \\
\textbf{Top-1-Accuracy:} 0.00 \\
\textbf{\SAS:} 0.55 \\
\textbf{Human Judgment:} 0.50
}}
  \caption{Exact match, F1-score, and top-1-accuracy are no good metrics to evaluate QA models. They do not take into account semantic similarity of predictions and ground-truth answers but only their lexical similarity. \SAS is close to human judgment of similarity.}\label{fig:placeholder2}%
\end{figure}

Extractive QA is not the only QA task that requires evaluation metrics that go beyond string-based matching.
Abstractive QA requires generative models to synthesize an answer and they need to be evaluated differently, too.
As of today, the evaluation of this task reuses metrics from the research field of natural language generation (NLG) but they are mostly string-based and not tailored to QA.
Given the shortcomings of the existing metrics, a novel metric for QA is needed and we address this challenge by presenting \SAS, a cross-encoder-based semantic answer similarity metric, and compare it with traditional lexical metrics and two recently proposed metrics for the more general task of semantic textual similarity (STS). 
To encourage and support research in this area, we release the annotated dataset\footnote{\url{https://semantic-answer-similarity.s3.amazonaws.com/data.zip}} and the trained model\footnote{\url{https://huggingface.co/deepset/gbert-large-sts}} under the Creative Commons Attribution-ShareAlike 4.0 International License (\href{https://creativecommons.org/licenses/by-sa/4.0/}{CC BY-SA 4.0}).

The remainder of this paper is structured as follows: Section~\ref{sec:related_work} describes related work on metrics used for the evaluation of QA and similar tasks.
In Section~\ref{sec:approach}, we present our new metric \SAS along with two existing STS metrics.
Three datasets, two newly created, three-way annotated datasets and one existing dataset from related work, are the basis of the experiments presented in Section~\ref{sec:experiments}, which compare four lexical and three semantic similarity metrics.
We conclude in Section~\ref{sec:conclusion} with directions for future work.

\section{Related Work}
\label{sec:related_work}
The related work discussed in this section goes beyond evaluation metrics explicitly meant for QA.
The task can be generalized to estimating the semantic similarity of a pair of text inputs, which is often referred to as semantic textual similarity (STS). 
While there is a benchmark dataset for STS estimation~\cite{cer2017semeval}, to the best of our knowledge, not even a single dataset has been created for the subtask of estimating semantic answer similarity. 
STS is closely related to paraphrasing, which, strictly speaking, refers to semantic equivalence.
As a consequence of this strict definition, \citet{bhagat-hovy-2013-squibs} introduce the concept of \emph{approximate} paraphrases as conveying a similar meaning.
Measuring to what extent a text is an approximate paraphrase of another text has been addressed in several subfields of research on natural language processing and we list different approaches in the following.

A recent tutorial summarizes evaluation metrics used in NLG, including but not limited to the QA task~\cite{khapra-sai-2021-tutorial}.
The challenge of evaluating NLG has also been recently addressed by~\citet{gehrmann-etal-2021-gem}, who introduce the GEM benchmark comprising eleven datasets.
Besides traditional, lexical similarity metrics, such as BLEU~\cite{papineni-etal-2002-bleu}, ROUGE~\cite{lin-hovy-2003-automatic}, and METEOR~\cite{banerjee2005meteor}, the benchmark also includes the two semantic similarity metrics BERTScore~\cite{Zhang2020BERTScore} and BLEURT~\cite{sellam-etal-2020-bleurt}.
Both BERTScore and BLEURT are BERT-based~\cite{Devlin.2019} metrics tailored to evaluating NLG.

BLEU (BiLingual Evaluation Understudy)~\cite{papineni-etal-2002-bleu} is a metric used to evaluate the quality of machine translations by measuring the n-gram overlap of prediction and ground truth.
Similarly, there is ROUGE (Recall-Oriented Understudy for Gisting Evaluation), which comes in different variations, such as using n-gram overlap (ROUGE-N)~\cite{lin-hovy-2003-automatic} or the longest common subsequence (ROUGE-L)~\cite{lin-och-2004-automatic} of prediction and ground truth.
METEOR~\cite{banerjee2005meteor} addresses the same task but aims to improve on BLEU, for example, by using a weighted harmonic mean of precision and recall of uni-gram overlap.
It relies on WordNet~\cite{miller1995wordnet} and, for this reason, can only be used for English.
There are also slightly modified versions of BLEU and Rouge for yes-no answers and entity answers that introduce a bonus term that gives more weight to correct answers of that type~\citet{yang-etal-2018-adaptations}.
Still, these modified versions rely on the standard implementations of BLEU and ROUGE, thus inheriting their shortcomings with regard to lexical vs.\ semantic similarity.

BERTScore~\cite{Zhang2020BERTScore} is similar to F1-score in that it performs stop word removal and lowercasing before the comparison.
TF-IDF is used to lower the influence of stop words on the score.
In our work, we argue that stopword removal should not be an extra step of the metric. 
Instead, the metrics should be based on models that have been trained to recognize what words and phrases are more or less important when comparing the meaning of two answers.
The main advantage of BERTScore over traditional metrics is that it compares contextual embeddings of tokens in the prediction and the ground truth instead of the actual tokens.
As future work, \citet{Zhang2020BERTScore} mention that BERTScore could be adapted for the evaluation of different tasks and in this paper, we discuss whether BERTScore is superior to other metrics for the evaluation of QA tasks.
\citet{chen-etal-2019-evaluating} apply BERTScore as an evaluation metric for QA and find that METEOR has a stronger correlation with human judgment.

While evaluating the correctness of predicted answers is the by far the most popular QA evaluation task, there are also approaches to evaluate the consistency of the predicted answers~\cite{ribeiro2019red} or other desirable properties of open-domain QA models, such as efficiency, context awareness, fine granularity of answers, end-to-end trainability or ability to generalize to different input data~\cite{ahmad2019reqa}.
Perturbations can serve as a means to evaluate the latter~\cite{shah-etal-2020-expect} and, in the same way, perturbations of the training data allow training models that are more robust~\cite{khashabi-etal-2020-bang}.
The correctness of answers can be estimated with methods from natural language inference: \citet{chen2021nli} convert answers to declarative statements and check whether the statement can be inferred from the relevant document (context).
\citet{nema-khapra-2018-towards} consider the task of evaluating the answerability of generated questions.
They find that existing string-based evaluation metrics do not correlate well with human judgment and propose modifications of these metrics, which give more weight to relevant content words, named entities, etc. 
With the metrics and the underlying models presented in this paper, we present end-to-end deep learning approaches, which learn these features automatically if they increase the correlation between the automated metric and human judgment.

Semantic similarity metrics might also mitigate the influence of an annotator bias on the evaluation, which has been reported to be learned by models and is currently not recognized if the same annotators create both the training and test dataset~\cite{geva-etal-2019-modeling,ko-etal-2020-look}.
That bias could be, for example, the position of the answer within a document always being in the first few sentences or a specific style of phrasing the questions.
With the help of semantic similarity metrics the position of the annotated answer within the context would not make a difference for the evaluation.

In line with the evaluation of QA models, the automated evaluation of models for question \emph{generation} also relies on BLEU, ROUGE, and METEOR, while human evaluation is limited to small datasets~\cite{du-etal-2017-learning}.
For the evaluation of conversational QA, \citet{siblini-etal-2021-towards} address the problems that arise from teacher forcing, which refers to earlier ground-truth answers being available to a model at each step in the conversation.
The authors discuss ideas to mitigate this problem, such as using the model's own predicted answers instead of the ground-truth answers. 
However, this approach only considers the ground-truth user reaction to the ground-truth answer but not the predicted answer as other reactions are not available in offline training and evaluation.
Last but not least, there is research on error analyses of QA models, which defines guidelines~\cite{wu2019errudite} or identifies challenges and promising directions for future work~\cite{rondeau-hazen-2018-systematic,wadhwa-etal-2018-comparative,pugaliya2019bend}.
These publications present anecdotal evidence of predictions that are evaluated as wrong due to the limitations of lexical similarity metrics but are in fact correct.
%
%
%
\section{Approach}
\label{sec:approach}
We consider four different approaches to estimate the semantic similarity of pairs of answers: a bi-encoder approach, a cross-encoder approach, a vanilla version of BERTScore, and a trained version of BERTScore.
This section describes each of these four approaches and the pre-trained language models they are based on.

\paragraph{Bi-Encoder}
The bi-encoder approach is based on the sentence transformers architecture~\cite{reimers-gurevych-2019-sentence}, which is a siamese neural network architecture comprising two language models that encode the two text inputs and cosine similarity to calculate a similarity score of the two encoded texts.
The model that we use is based on xlm-roberta-base, where the training has been continued on an unreleased multi-lingual paraphrase dataset.
The resulting model, called 
paraphrase-xlm-r-multilingual-v1, has then been fine-tuned on the English-language STS benchmark dataset~\cite{cer2017semeval} and a machine-translated German-language version\footnote{\url{https://github.com/t-systems-on-site-services-gmbh/german-STSbenchmark}} of the same data.
The final model is called T-Systems-onsite/cross-en-de-roberta-sentence-transformer and is available on the huggingface model hub.
As the model has been trained on English- and German-language data, we use the exact same model for all three datasets in our experiments.
An advantage of the bi-encoder architecture is that the embeddings of the two text inputs are calculated separately.
As a consequence, the embeddings of the ground-truth answers can be pre-computed and reused when comparing with the predictions of several different models. 
This pre-computation can almost halve the time needed to run the evaluation.


\paragraph{SAS}
Our new approach called SAS differs from the bi-encoder in that it does not calculate separate embeddings for the input texts. Instead, we use a cross-encoder architecture, where the two texts are concatenated with a special separator token in between.
The underlying language model is called cross-encoder/stsb-roberta-large and has been trained on the STS benchmark dataset~\cite{cer2017semeval}.
Unfortunately, there are only English cross-encoder models for STS estimation available.
Therefore, we train a German cross-encoder model for STS estimation, which we release online.
This model is based on deepset/gbert-large and we train it for four epochs with a batch size of 16 and the Adam optimizer on the machine-translated version of the STS benchmark that has previously been used to train a bi-encoder STS model for German.
For the warm-up phase of the training, we use 10\% of the training data and linearly increase the learning rate to 2e-5.
While pre-computation is not possible with the cross-encoder architecture, its advantage over bi-encoders is that it takes into account both text inputs at the same time when applying the language model in a monolithic way rather than calculating encodings separately and comparing them afterward.


\paragraph{BERTScore vanilla or trained}
The BERTScore vanilla approach uses the task-agnostic, pre-trained language models bert-base-uncased for the English-language datasets and deepset/gelectra-base for the German-language dataset. 
In line with the approach by \citet{Zhang2020BERTScore}, we use the language models to generate contextual embeddings, match the embeddings of the tokens in the ground-truth answer and in the prediction and take the maximum cosine similarity of the matched tokens as the similarity score of the two answers. 
The optional step of importance weighting of tokens based on inverse document frequency scores is not applied.
For the vanilla version, we extract embeddings from the second layer and for the trained version from the last layer.
%
%
In contrast to the vanilla model, the BERTScore trained model uses a task-specific model tailored to STS estimation.
It is the same multi-lingual model that is used by the bi-encoder approach, called T-Systems-onsite/cross-en-de-roberta-sentence-transformer.

\section{Experiments}
\label{sec:experiments}
To evaluate the ability of the different approaches to estimate semantic answer similarity, we measure their correlation with human judgment of similarity on three datasets.
This section describes the dataset creation, experiment setup, and the final results.

\subsection{Datasets}
The evaluation uses subsets of three existing datasets: SQuAD, GermanQuAD, and NQ-open.
We process and hand-annotate the datasets as described in the following so that each of the processed subsets contains pairs of answers and a class label that indicates their semantic similarity.
There are three similarity classes: dissimilar answers, approximately similar answers, and equivalent answers, which are all described in Table~\ref{table:classes}.

\begin{table}
\centering
\begin{tabular}{lp{6.6cm}}
\toprule
\multirow{2}{*}{0} & The two answers are completely dissimilar. \\
& ``power steering'' $\neq$ ``air conditioning''\\
%
\midrule
\multirow{4}{*}{1} & The two answers have a similar meaning but one of them is less detailed and could be derived from the more elaborate answer. \\
& ``Joseph Priestley'' $\approx$ ``Priestley''\\
\midrule
\multirow{2}{*}{2} & The two answers have the same meaning. \\
& ``UV'' $=$ ``ultraviolet''\\
\bottomrule
\end{tabular}
\caption{Similarity scores with descriptions and example pairs of answers.}
\label{table:classes}
\end{table}

\paragraph{SQuAD}
We annotate the semantic similarity of pairs of answers in a subset of the English-language SQuAD test dataset~\cite{rajpurkar2018know}.
The original dataset contains multi-way annotated questions, which means there are on average 4.8 answers per question.
Answers to the same question by different annotators often are the same but in some cases they have only a small overlap or no overlap at all.
We consider a subset where 566 pairs of ground-truth answers have an F1-score of 0 (no lexical overlap of the answers) and 376 pairs have an F1-score larger than 0 (some lexical overlap of the answers).
As we use the majority vote as the ground-truth label of semantic similarity in our experiments, we let two of the authors label each pair of answers while a third author acts as a tie-breaker labeling only those samples, where the first two labels disagree.
The resulting dataset comprises 942 pairs of answers each with a majority vote indicating either dissimilar answers, approximately similar answers, or equivalent answers.

\paragraph{GermanQuAD}
To show that the presented approaches also work on non-English datasets, we consider the German-language GermanQuAD dataset~\cite{moeller2021germanquad}.
It contains a three-way annotated test set, which means there are three correct answers given for each question. 
After removing questions where all answers are the same, there are 137 pairs of ground-truth answers that have an F1-score of 0 and 288 pairs of answers have an F1-score larger than 0.
We label these 425 pairs in the same way as the SQuAD subset resulting in 425 pairs of answers each with a majority vote indicating their semantic similarity.

\paragraph{NQ-open} The original Natural Questions dataset (NQ)~\cite{Kwiatkowski.2019} was meant for reading comprehension but \citet{lee-etal-2019-latent} adapted the dataset for open-domain QA and it has been released under the name NQ-open.
We use the test dataset of NQ-open as it contains not only questions and ground-truth answers but also model predictions and annotations how similar these predictions are to the ground-truth answer.
There are three classes of definitely incorrect predictions, possibly correct predictions, and definitely correct predictions.
\citet{min2021neurips} report in more detail on how these additional annotations were created.
They resemble the three similarity classes we defined in Table~\ref{table:classes}.
After filtering for only those questions that have exactly one ground-truth answer, we create pairs of ground-truth answers and model predictions accompanied with the label indicating the correctness of the prediction, which also corresponds to the similarity of the ground-truth answer and the predicted answer.
There are 3,658 pairs of answers of which 3118 have an F1-score of 0 and 540 pairs that have an F1-score larger than 0.

\subsection{Results}
\begin{table*}
\setlength{\tabcolsep}{5pt}
\centering
\begin{tabular}{lcccccccccccc}
    \toprule
    & \multicolumn{4}{c}{\textbf{SQuAD}} & \multicolumn{4}{c}{\textbf{GermanQuAD}} & \multicolumn{4}{c}{\textbf{NQ-open}} \\
    \cmidrule(lr){2-5} \cmidrule(lr){6-9} \cmidrule(lr){10-13}
     &
    \multicolumn{2}{c}{\bm{$F1=0$}} & \multicolumn{2}{c}{\bm{$F1\neq 0$}} & \multicolumn{2}{c}{\bm{$F1=0$}} & \multicolumn{2}{c}{\bm{$F1\neq 0$}} & \multicolumn{2}{c}{\bm{$F1=0$}} & \multicolumn{2}{c}{\bm{$F1\neq 0$}}\\
    \cmidrule(lr){2-3}  \cmidrule(lr){4-5} \cmidrule(lr){6-7} \cmidrule(lr){8-9} \cmidrule(lr){10-11} \cmidrule(lr){12-13} 
    \textbf{Metrics} & $\bm{r}$ & $\bm{\tau}$ &$\bm{r}$ & $\bm{\tau}$ & $\bm{r}$ & $\bm{\tau}$ &$\bm{r}$ & $\bm{\tau}$ & $\bm{r}$ & $\bm{\tau}$ &$\bm{r}$ & $\bm{\tau}$\\
    \midrule
     Human             & 0.61 & 0.48 & 0.68 & 0.64 & 0.64 & 0.57 & 0.57 & 0.54 & -    & -    & -    &  -  \\
     \midrule
     BLEU              &  00.0   &  00.0   & 0.18 & 0.16 & 00.0    & 00.0    & 0.13 & 0.05 & 00.0    & 00.0    & 0.08 & 0.08\\
     ROUGE-L           & 0.10 & 0.04 & 0.56 & 0.46 & 0.16 & 0.02 & 0.54 & 0.43 & 0.14 & 0.12 & 0.40 & 0.33\\
     METEOR            & 0.38 & 0.19 & 0.45 & 0.37 & -    &    - &  -   & -    & 0.21 & 0.16 & 0.30 & 0.25\\
     F1-score          & 00.0 &   00.0  & 0.60 & 0.50 & 00.0    & 00.0    & 0.55& 0.44 &   00.0  &   00.0  & 0.37 & 0.31\\
     Bi-Encoder & 0.48 & 0.30 & 0.69 & 0.57 & 0.39 & 0.27 & 0.56 &  0.47& 0.23 & 0.15 & 0.38 & 0.30 \\
     BERTScore vanilla & 0.27 & 0.15 & 0.61 & 0.48 & 0.21 & 0.01 & 0.52 & 0.41 & 0.14 & 0.13 & 0.16 & 0.11\\
     BERTScore trained & 0.52 &\B0.32& 0.70 & 0.57 &  0.41&  0.28& 0.57 & 0.47 &0.25 &\B0.16& 0.38 & 0.30 \\
     \SAS (ours)     &\B0.56 & 0.29&\B0.75&\B0.61&\B0.49&\B0.33&\B0.68&\B0.55&\B0.31& 0.13 &\B0.54&\B0.42\\
    \bottomrule
\end{tabular}
\caption{Correlation between human judgment and automated metrics using Pearson's ($r$) and Kendall's tau-b ($\tau$) rank correlation coefficients on subsets of SQuAD, GermanQuAD, and NQ-open. Human baseline correlations on SQuAD and GermanQUAD were measured between first and second annotator. METEOR is not available for German and scores of individual annotators have not been reported on NQ-open (indicated as ``-'' in the table). }
\label{table:correlation}
\end{table*}
Table~\ref{table:correlation} lists the correlation between different automated evaluation metrics and human judgment using Pearson's r and Kendall's tau-b rank correlation coefficients on labeled subsets of SQuAD, GermanQuAD, and NQ-open datasets.
The traditional metrics ROUGE-L and METEOR have very weak correlation with human judgement if there is no lexical overlap between the pair of answers, in which case the F1-score and BLEU are 0.
If there is some lexical overlap, the correlation is stronger for all these metrics but BLEU lags far behind the others.
METEOR is outperformed by ROUGE-L and F1-score, which achieve almost equal correlation.
All four semantic answer similarity approaches outperform the traditional metrics and among them, the cross-encoder model is consistently achieving the strongest correlation with human judgment except for slightly underperforming the trained BERTScore metric with regard to $\tau$ on English-language pairs of answers with no lexical overlap. 
This result shows that semantic similarity metrics are needed in addition to lexical-based metrics for automated evaluation of QA models. 
The former correlate much better with human judgment and thus, are a better estimation of a model's performance in real-world applications.





\begin{figure}[!htb]
\centering
\includegraphics[width=1.0\linewidth]{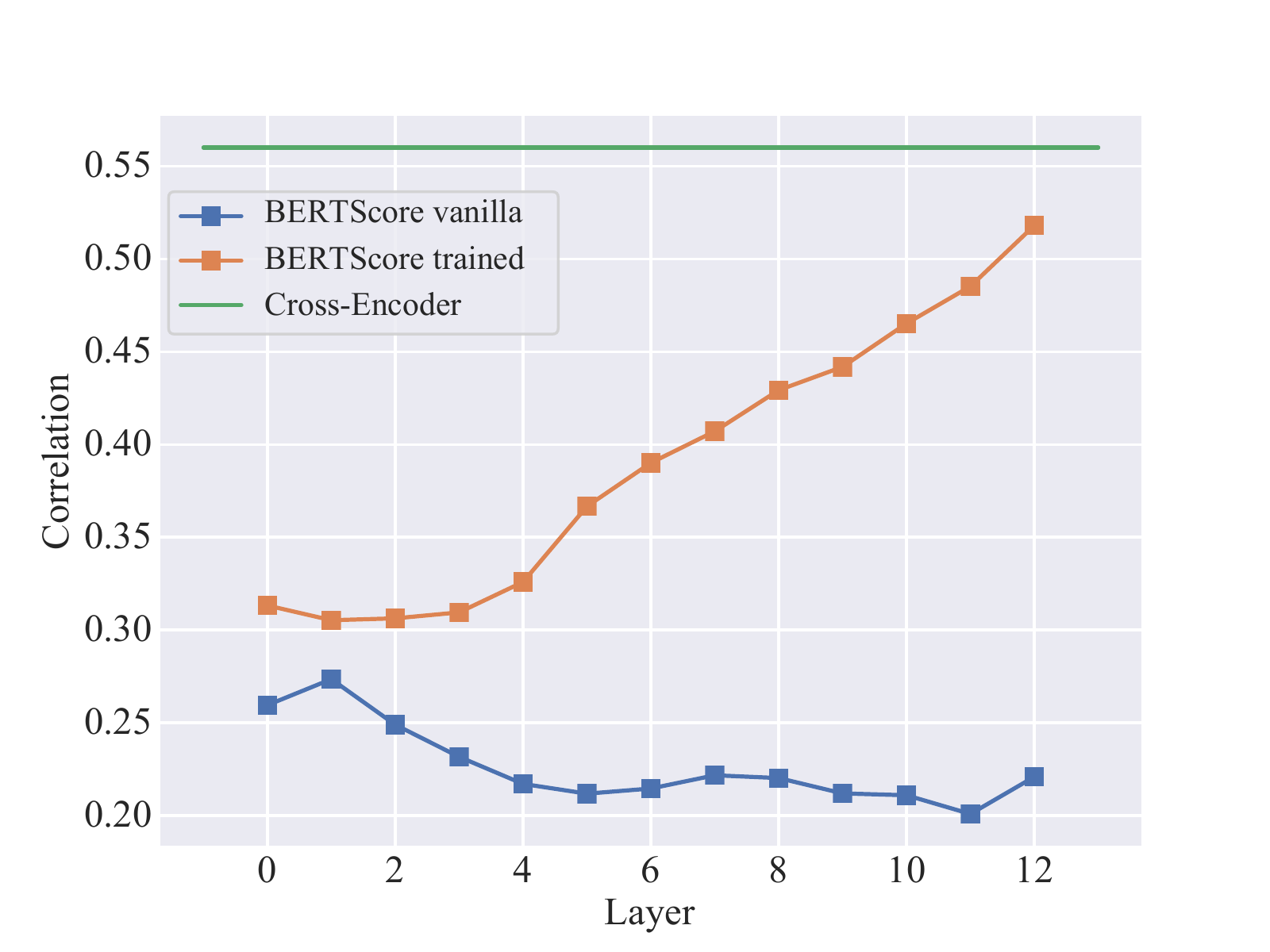}
\caption{Pearson correlation between BERTScore (computed across different layers) and human judgement of similarity of answer pairs on SQuAD dev set. BERTScore vanilla is pretrained only on Wikipedia, whereas BERTScore trained is fine-tuned on the STS benchmark dataset~\cite{cer2017semeval}.}
\label{fig:num_layers}
\end{figure}

\paragraph{Embedding Extraction for BERTScore}
BERT\-Score can be used with different language models to generate contextual embeddings of text inputs.
While the embeddings are typically extracted from the last layer of the model, they can be extracted from any of its layers and related work has shown that for some tasks the last layer is not the best~\cite{liu-etal-2019-linguistic}.
The experiment visualized in Figure~\ref{fig:num_layers} evaluates the correlation between human judgment of semantic answer similarity and a vanilla and a trained BERTScore model.
Comparing the extraction of embeddings from the different layers, we find that the last layer drastically outperforms all other models for the trained model.
For the vanilla BERTScore model, the choice of the layer has a much smaller influence on the performance, with the first two layers resulting in the strongest correlation with human judgment.
For comparison, Figure~\ref{fig:num_layers} also includes the results of a cross-encoder model, which does not have the option to choose different layers due to its architecture.

\section{Conclusion and Future Work}
\label{sec:conclusion}
Current evaluation metrics for QA models are limited in that they check for lexical or positional overlap of ground-truth answers and predictions but do not take into account semantic similarity.
In this paper, we present \SAS, a semantic answer similarity metric that overcomes this limitation.
It leverages a cross-encoder architecture and transformer-based language models, which are pre-trained on STS datasets that have not been used in the context of QA so far.
Experiments on three datasets demonstrate that \SAS outperforms four lexical-based and three semantics-based similarity metrics regarding the correlation between the automated metrics and human judgment of the semantic similarity of pairs of answers.
A promising path for future work is to analyze pairs of answers where \SAS differs from human judgment and find types of common errors.
Based on these findings, a dataset tailored to training models for estimating semantic answer similarity could be created.


%


\bibliography{custom}

\begin{thebibliography}{35}
\expandafter\ifx\csname natexlab\endcsname\relax\def\natexlab#1{#1}\fi

\bibitem[{Ahmad et~al.(2019)Ahmad, Constant, Yang, and Cer}]{ahmad2019reqa}
Amin Ahmad, Noah Constant, Yinfei Yang, and Daniel Cer. 2019.
\newblock \href {https://aclanthology.org/D19-5819/} {Reqa: An evaluation for
  end-to-end answer retrieval models}.
\newblock In \emph{Proceedings of the Workshop on Machine Reading for Question
  Answering (MRQA@EMNLP-IJCNLP)}, pages 137--146. Association for Computational
  Linguistics.

\bibitem[{Banerjee and Lavie(2005)}]{banerjee2005meteor}
Satanjeev Banerjee and Alon Lavie. 2005.
\newblock \href {https://aclanthology.org/W05-0909/} {Meteor: An automatic
  metric for mt evaluation with improved correlation with human judgments}.
\newblock In \emph{Proceedings of the ACL Workshop on Intrinsic and Extrinsic
  Evaluation Measures for Machine Translation and/or Summarization}, pages
  65--72. Association for Computational Linguistics.

\bibitem[{Bhagat and Hovy(2013)}]{bhagat-hovy-2013-squibs}
Rahul Bhagat and Eduard Hovy. 2013.
\newblock \href {https://doi.org/10.1162/COLI_a_00166} {{S}quibs: What is a
  paraphrase?}
\newblock \emph{Computational Linguistics}, 39(3):463--472.

\bibitem[{Cer et~al.(2017)Cer, Diab, Agirre, Lopez-Gazpio, and
  Specia}]{cer2017semeval}
Daniel Cer, Mona Diab, Eneko Agirre, Inigo Lopez-Gazpio, and Lucia Specia.
  2017.
\newblock \href {https://arxiv.org/abs/1708.00055} {Semeval-2017 task 1:
  Semantic textual similarity-multilingual and cross-lingual focused
  evaluation}.
\newblock In \emph{Proceedings of the International Workshop on Semantic
  Evaluations (SemEval@ACL}, pages 1--14. Association for Computational
  Linguistics.

\bibitem[{Chen et~al.(2019)Chen, Stanovsky, Singh, and
  Gardner}]{chen-etal-2019-evaluating}
Anthony Chen, Gabriel Stanovsky, Sameer Singh, and Matt Gardner. 2019.
\newblock \href {https://doi.org/10.18653/v1/D19-5817} {Evaluating question
  answering evaluation}.
\newblock In \emph{Proceedings of the Workshop on Machine Reading for Question
  Answering (MRQA@EMNLP-IJCNLP)}, pages 119--124. Association for Computational
  Linguistics.

\bibitem[{Chen et~al.(2021)Chen, Choi, and Durrett}]{chen2021nli}
Jifan Chen, Eunsol Choi, and Greg Durrett. 2021.
\newblock \href {https://arxiv.org/abs/2104.08731} {Can {NLI} models verify
  {QA} systems' predictions?}
\newblock \emph{arXiv preprint arXiv:2104.08731}.

\bibitem[{Devlin et~al.(2019)Devlin, Chang, Lee, and Toutanova}]{Devlin.2019}
Jacob Devlin, Ming-Wei Chang, Kenton Lee, and Kristina Toutanova. 2019.
\newblock \href {https://doi.org/10.18653/v1/N19-1423} {{BERT}: Pre-training of
  deep bidirectional transformers for language understanding}.
\newblock In \emph{Proceedings of the Conference of the North {A}merican
  Chapter of the Association for Computational Linguistics: Human Language
  Technologies (NAACL-HLT)}, pages 4171--4186. Association for Computational
  Linguistics.

\bibitem[{Du et~al.(2017)Du, Shao, and Cardie}]{du-etal-2017-learning}
Xinya Du, Junru Shao, and Claire Cardie. 2017.
\newblock \href {https://doi.org/10.18653/v1/P17-1123} {Learning to ask: Neural
  question generation for reading comprehension}.
\newblock In \emph{Proceedings of the Annual Meeting of the Association for
  Computational Linguistics (ACL)}, pages 1342--1352. Association for
  Computational Linguistics.

\bibitem[{Gehrmann et~al.(2021)Gehrmann, Adewumi, Aggarwal, Ammanamanchi,
  Aremu, Bosselut, Chandu, Clinciu, Das, Dhole, Du, Durmus, Du{\v{s}}ek,
  Emezue, Gangal, Garbacea, Hashimoto, Hou, Jernite, Jhamtani, Ji, Jolly, Kale,
  Kumar, Ladhak, Madaan, Maddela, Mahajan, Mahamood, Majumder, Martins,
  McMillan-Major, Mille, van Miltenburg, Nadeem, Narayan, Nikolaev,
  Niyongabo~Rubungo, Osei, Parikh, Perez-Beltrachini, Rao, Raunak, Rodriguez,
  Santhanam, Sedoc, Sellam, Shaikh, Shimorina, Sobrevilla~Cabezudo, Strobelt,
  Subramani, Xu, Yang, Yerukola, and Zhou}]{gehrmann-etal-2021-gem}
Sebastian Gehrmann, Tosin Adewumi, Karmanya Aggarwal, Pawan~Sasanka
  Ammanamanchi, Anuoluwapo Aremu, Antoine Bosselut, Khyathi~Raghavi Chandu,
  Miruna-Adriana Clinciu, Dipanjan Das, Kaustubh Dhole, Wanyu Du, Esin Durmus,
  Ond{\v{r}}ej Du{\v{s}}ek, Chris~Chinenye Emezue, Varun Gangal, Cristina
  Garbacea, Tatsunori Hashimoto, Yufang Hou, Yacine Jernite, Harsh Jhamtani,
  Yangfeng Ji, Shailza Jolly, Mihir Kale, Dhruv Kumar, Faisal Ladhak, Aman
  Madaan, Mounica Maddela, Khyati Mahajan, Saad Mahamood, Bodhisattwa~Prasad
  Majumder, Pedro~Henrique Martins, Angelina McMillan-Major, Simon Mille, Emiel
  van Miltenburg, Moin Nadeem, Shashi Narayan, Vitaly Nikolaev, Andre
  Niyongabo~Rubungo, Salomey Osei, Ankur Parikh, Laura Perez-Beltrachini,
  Niranjan~Ramesh Rao, Vikas Raunak, Juan~Diego Rodriguez, Sashank Santhanam,
  Jo{\~a}o Sedoc, Thibault Sellam, Samira Shaikh, Anastasia Shimorina,
  Marco~Antonio Sobrevilla~Cabezudo, Hendrik Strobelt, Nishant Subramani, Wei
  Xu, Diyi Yang, Akhila Yerukola, and Jiawei Zhou. 2021.
\newblock \href {https://doi.org/10.18653/v1/2021.gem-1.10} {The {GEM}
  benchmark: Natural language generation, its evaluation and metrics}.
\newblock In \emph{Proceedings of the Workshop on Natural Language Generation,
  Evaluation, and Metrics (GEM@ACL-IJCNLP)}, pages 96--120. Association for
  Computational Linguistics.

\bibitem[{Geva et~al.(2019)Geva, Goldberg, and
  Berant}]{geva-etal-2019-modeling}
Mor Geva, Yoav Goldberg, and Jonathan Berant. 2019.
\newblock \href {https://doi.org/10.18653/v1/D19-1107} {Are we modeling the
  task or the annotator? an investigation of annotator bias in natural language
  understanding datasets}.
\newblock In \emph{Proceedings of the Conference on Empirical Methods in
  Natural Language Processing and the International Joint Conference on Natural
  Language Processing (EMNLP-IJCNLP)}, pages 1161--1166. Association for
  Computational Linguistics.

\bibitem[{Khapra and Sai(2021)}]{khapra-sai-2021-tutorial}
Mitesh~M. Khapra and Ananya~B. Sai. 2021.
\newblock \href {https://doi.org/10.18653/v1/2021.naacl-tutorials.4} {A
  tutorial on evaluation metrics used in natural language generation}.
\newblock In \emph{Proceedings of the Conference of the North American Chapter
  of the Association for Computational Linguistics: Human Language Technologies
  (NAACL-HLT)}, pages 15--19. Association for Computational Linguistics.

\bibitem[{Khashabi et~al.(2020)Khashabi, Khot, and
  Sabharwal}]{khashabi-etal-2020-bang}
Daniel Khashabi, Tushar Khot, and Ashish Sabharwal. 2020.
\newblock \href {https://doi.org/10.18653/v1/2020.emnlp-main.12} {More bang for
  your buck: Natural perturbation for robust question answering}.
\newblock In \emph{Proceedings of the Conference on Empirical Methods in
  Natural Language Processing (EMNLP)}, pages 163--170. Association for
  Computational Linguistics.

\bibitem[{Ko et~al.(2020)Ko, Lee, Kim, Kim, and Kang}]{ko-etal-2020-look}
Miyoung Ko, Jinhyuk Lee, Hyunjae Kim, Gangwoo Kim, and Jaewoo Kang. 2020.
\newblock \href {https://doi.org/10.18653/v1/2020.emnlp-main.84} {Look at the
  first sentence: Position bias in question answering}.
\newblock In \emph{Proceedings of the Conference on Empirical Methods in
  Natural Language Processing (EMNLP)}, pages 1109--1121. Association for
  Computational Linguistics.

\bibitem[{Kwiatkowski et~al.(2019)Kwiatkowski, Palomaki, Redfield, Collins,
  Parikh, Alberti, Epstein, Polosukhin, Devlin, Lee, Toutanova, Jones, Kelcey,
  Chang, Dai, Uszkoreit, Le, and Petrov}]{Kwiatkowski.2019}
Tom Kwiatkowski, Jennimaria Palomaki, Olivia Redfield, Michael Collins, Ankur
  Parikh, Chris Alberti, Danielle Epstein, Illia Polosukhin, Jacob Devlin,
  Kenton Lee, Kristina Toutanova, Llion Jones, Matthew Kelcey, Ming-Wei Chang,
  Andrew~M. Dai, Jakob Uszkoreit, Quoc Le, and Slav Petrov. 2019.
\newblock \href {https://doi.org/10.1162/tacl_a_00276} {Natural questions: A
  benchmark for question answering research}.
\newblock \emph{Transactions of the Association for Computational Linguistics
  (TACL)}, 7:452--466.

\bibitem[{Lee et~al.(2019)Lee, Chang, and Toutanova}]{lee-etal-2019-latent}
Kenton Lee, Ming-Wei Chang, and Kristina Toutanova. 2019.
\newblock \href {https://doi.org/10.18653/v1/P19-1612} {Latent retrieval for
  weakly supervised open domain question answering}.
\newblock In \emph{Proceedings of the Annual Meeting of the Association for
  Computational Linguistics (ACL)}, pages 6086--6096. Association for
  Computational Linguistics.

\bibitem[{Lin and Hovy(2003)}]{lin-hovy-2003-automatic}
Chin-Yew Lin and Eduard Hovy. 2003.
\newblock \href {https://aclanthology.org/N03-1020} {Automatic evaluation of
  summaries using n-gram co-occurrence statistics}.
\newblock In \emph{Proceedings of the Human Language Technology Conference of
  the North {A}merican Chapter of the Association for Computational Linguistics
  (NAACL-HLT)}, pages 150--157.

\bibitem[{Lin and Och(2004)}]{lin-och-2004-automatic}
Chin-Yew Lin and Franz~Josef Och. 2004.
\newblock \href {https://doi.org/10.3115/1218955.1219032} {Automatic evaluation
  of machine translation quality using longest common subsequence and
  skip-bigram statistics}.
\newblock In \emph{Proceedings of the Annual Meeting of the Association for
  Computational Linguistics (ACL)}, pages 605--612.

\bibitem[{Liu et~al.(2019)Liu, Gardner, Belinkov, Peters, and
  Smith}]{liu-etal-2019-linguistic}
Nelson~F. Liu, Matt Gardner, Yonatan Belinkov, Matthew~E. Peters, and Noah~A.
  Smith. 2019.
\newblock \href {https://doi.org/10.18653/v1/N19-1112} {Linguistic knowledge
  and transferability of contextual representations}.
\newblock In \emph{Proceedings of the Conference of the North {A}merican
  Chapter of the Association for Computational Linguistics: Human Language
  Technologies (NAACL-HLT)}, pages 1073--1094. Association for Computational
  Linguistics.

\bibitem[{Miller(1995)}]{miller1995wordnet}
George~A Miller. 1995.
\newblock Wordnet: a lexical database for english.
\newblock \emph{Communications of the ACM}, 38(11):39--41.

\bibitem[{Min et~al.(2021)Min, Boyd-Graber, Alberti, Chen, Choi, Collins, Guu,
  Hajishirzi, Lee, Palomaki, Raffel, Roberts, Kwiatkowski, Lewis, Wu, Küttler,
  Liu, Minervini, Stenetorp, Riedel, Yang, Seo, Izacard, Petroni, Hosseini,
  Cao, Grave, Yamada, Shimaoka, Suzuki, Miyawaki, Sato, Takahashi, Suzuki,
  Fajcik, Docekal, Ondrej, Smrz, Cheng, Shen, Liu, He, Chen, Gao, Oguz, Chen,
  Karpukhin, Peshterliev, Okhonko, Schlichtkrull, Gupta, Mehdad, and tau
  Yih}]{min2021neurips}
Sewon Min, Jordan Boyd-Graber, Chris Alberti, Danqi Chen, Eunsol Choi, Michael
  Collins, Kelvin Guu, Hannaneh Hajishirzi, Kenton Lee, Jennimaria Palomaki,
  Colin Raffel, Adam Roberts, Tom Kwiatkowski, Patrick Lewis, Yuxiang Wu,
  Heinrich Küttler, Linqing Liu, Pasquale Minervini, Pontus Stenetorp,
  Sebastian Riedel, Sohee Yang, Minjoon Seo, Gautier Izacard, Fabio Petroni,
  Lucas Hosseini, Nicola~De Cao, Edouard Grave, Ikuya Yamada, Sonse Shimaoka,
  Masatoshi Suzuki, Shumpei Miyawaki, Shun Sato, Ryo Takahashi, Jun Suzuki,
  Martin Fajcik, Martin Docekal, Karel Ondrej, Pavel Smrz, Hao Cheng, Yelong
  Shen, Xiaodong Liu, Pengcheng He, Weizhu Chen, Jianfeng Gao, Barlas Oguz,
  Xilun Chen, Vladimir Karpukhin, Stan Peshterliev, Dmytro Okhonko, Michael
  Schlichtkrull, Sonal Gupta, Yashar Mehdad, and Wen tau Yih. 2021.
\newblock \href {https://arxiv.org/abs/2101.00133} {{NeurIPS} 2020
  {EfficientQA} competition: Systems, analyses and lessons learned}.
\newblock \emph{arXiv preprint arXiv:2101.00133}.

\bibitem[{Möller et~al.(2021)Möller, Risch, and
  Pietsch}]{moeller2021germanquad}
Timo Möller, Julian Risch, and Malte Pietsch. 2021.
\newblock \href {https://arxiv.org/abs/2104.12741} {{GermanQuAD} and
  {GermanDPR}: Improving non-english question answering and passage retrieval}.
\newblock \emph{arXiv preprint arXiv:2104.12741}.

\bibitem[{Nema and Khapra(2018)}]{nema-khapra-2018-towards}
Preksha Nema and Mitesh~M. Khapra. 2018.
\newblock \href {https://doi.org/10.18653/v1/D18-1429} {Towards a better metric
  for evaluating question generation systems}.
\newblock In \emph{Proceedings of the Conference on Empirical Methods in
  Natural Language Processing (EMNLP)}, pages 3950--3959. Association for
  Computational Linguistics.

\bibitem[{Papineni et~al.(2002)Papineni, Roukos, Ward, and
  Zhu}]{papineni-etal-2002-bleu}
Kishore Papineni, Salim Roukos, Todd Ward, and Wei-Jing Zhu. 2002.
\newblock \href {https://doi.org/10.3115/1073083.1073135} {{B}leu: a method for
  automatic evaluation of machine translation}.
\newblock In \emph{Proceedings of the Annual Meeting of the Association for
  Computational Linguistics (ACL)}, pages 311--318. Association for
  Computational Linguistics.

\bibitem[{Pugaliya et~al.(2019)Pugaliya, Route, Ma, Geng, and
  Nyberg}]{pugaliya2019bend}
Hemant Pugaliya, James Route, Kaixin Ma, Yixuan Geng, and Eric Nyberg. 2019.
\newblock \href {https://aclanthology.org/D19-5818.pdf} {Bend but don’t
  break? multi-challenge stress test for qa models}.
\newblock In \emph{Proceedings of the Workshop on Machine Reading for Question
  Answering (MRQA@EMNLP-IJCNLP)}, pages 125--136. Association for Computational
  Linguistics.

\bibitem[{Rajpurkar et~al.(2018)Rajpurkar, Jia, and Liang}]{rajpurkar2018know}
Pranav Rajpurkar, Robin Jia, and Percy Liang. 2018.
\newblock \href {https://www.aclweb.org/anthology/P18-2124} {Know what you
  don{'}t know: Unanswerable questions for {SQ}u{AD}}.
\newblock In \emph{Proceedings of the Annual Meeting of the Association for
  Computational Linguistics (ACL)}, pages 784--789. Association for
  Computational Linguistics.

\bibitem[{Reimers and Gurevych(2019)}]{reimers-gurevych-2019-sentence}
Nils Reimers and Iryna Gurevych. 2019.
\newblock \href {https://doi.org/10.18653/v1/D19-1410} {Sentence-{BERT}:
  Sentence embeddings using {S}iamese {BERT}-networks}.
\newblock In \emph{Proceedings of the Conference on Empirical Methods in
  Natural Language Processing and the 9th International Joint Conference on
  Natural Language Processing (EMNLP-IJCNLP)}, pages 3982--3992. Association
  for Computational Linguistics.

\bibitem[{Ribeiro et~al.(2019)Ribeiro, Guestrin, and Singh}]{ribeiro2019red}
Marco~Tulio Ribeiro, Carlos Guestrin, and Sameer Singh. 2019.
\newblock \href {https://aclanthology.org/P19-1621.pdf} {Are red roses red?
  evaluating consistency of question-answering models}.
\newblock In \emph{Proceedings of the Annual Meeting of the Association for
  Computational Linguistics (ACL)}, pages 6174--6184. Association for
  Computational Linguistics.

\bibitem[{Rondeau and Hazen(2018)}]{rondeau-hazen-2018-systematic}
Marc-Antoine Rondeau and T.~J. Hazen. 2018.
\newblock \href {https://doi.org/10.18653/v1/W18-2602} {Systematic error
  analysis of the {S}tanford question answering dataset}.
\newblock In \emph{Proceedings of the Workshop on Machine Reading for Question
  Answering (MRQA@ACL)}, pages 12--20. Association for Computational
  Linguistics.

\bibitem[{Sellam et~al.(2020)Sellam, Das, and Parikh}]{sellam-etal-2020-bleurt}
Thibault Sellam, Dipanjan Das, and Ankur Parikh. 2020.
\newblock \href {https://doi.org/10.18653/v1/2020.acl-main.704} {{BLEURT}:
  Learning robust metrics for text generation}.
\newblock In \emph{Proceedings of the Annual Meeting of the Association for
  Computational Linguistics (ACL)}, pages 7881--7892. Association for
  Computational Linguistics.

\bibitem[{Shah et~al.(2020)Shah, Gupta, and Roth}]{shah-etal-2020-expect}
Krunal Shah, Nitish Gupta, and Dan Roth. 2020.
\newblock \href {https://doi.org/10.18653/v1/2020.findings-emnlp.317} {What do
  we expect from multiple-choice {QA} systems?}
\newblock In \emph{Findings of the Association for Computational Linguistics:
  EMNLP}, pages 3547--3553. Association for Computational Linguistics.

\bibitem[{Siblini et~al.(2021)Siblini, Sayil, and
  Kessaci}]{siblini-etal-2021-towards}
Wissam Siblini, Baris Sayil, and Yacine Kessaci. 2021.
\newblock \href {https://aclanthology.org/2021.acl-short.130/} {Towards a more
  robust evaluation for conversational question answering}.
\newblock In \emph{Proceedings of the Annual Meeting of the Association for
  Computational Linguistics and the International Joint Conference on Natural
  Language Processing (ACL-IJCNLP)}, pages 1028--1034. Association for
  Computational Linguistics.

\bibitem[{Wadhwa et~al.(2018)Wadhwa, Chandu, and
  Nyberg}]{wadhwa-etal-2018-comparative}
Soumya Wadhwa, Khyathi Chandu, and Eric Nyberg. 2018.
\newblock \href {https://doi.org/10.18653/v1/W18-2610} {Comparative analysis of
  neural {QA} models on {SQ}u{AD}}.
\newblock In \emph{Proceedings of the Workshop on Machine Reading for Question
  Answering (MRQA@ACL)}, pages 89--97. Association for Computational
  Linguistics.

\bibitem[{Wu et~al.(2019)Wu, Ribeiro, Heer, and Weld}]{wu2019errudite}
Tongshuang Wu, Marco~Tulio Ribeiro, Jeffrey Heer, and Daniel~S Weld. 2019.
\newblock \href {https://aclanthology.org/P19-1073.pdf} {Errudite: Scalable,
  reproducible, and testable error analysis}.
\newblock In \emph{Proceedings of the Annual Meeting of the Association for
  Computational Linguistics (ACL)}, pages 747--763. Association for
  Computational Linguistics.

\bibitem[{Yang et~al.(2018)Yang, Liu, Liu, Lyu, and
  Li}]{yang-etal-2018-adaptations}
An~Yang, Kai Liu, Jing Liu, Yajuan Lyu, and Sujian Li. 2018.
\newblock \href {https://doi.org/10.18653/v1/W18-2611} {Adaptations of {ROUGE}
  and {BLEU} to better evaluate machine reading comprehension task}.
\newblock In \emph{Proceedings of the Workshop on Machine Reading for Question
  Answering (MRQA@ACL)}, pages 98--104. Association for Computational
  Linguistics.

\bibitem[{Zhang et~al.(2020)Zhang, Kishore, Wu, Weinberger, and
  Artzi}]{Zhang2020BERTScore}
Tianyi Zhang, Varsha Kishore, Felix Wu, Kilian~Q. Weinberger, and Yoav Artzi.
  2020.
\newblock \href {https://openreview.net/forum?id=SkeHuCVFDr} {{BERTScore}:
  Evaluating text generation with bert}.
\newblock In \emph{International Conference on Learning Representations
  (ICLR)}.

\end{thebibliography}
\bibliographystyle{acl_natbib}

\end{document}